\documentclass[runningheads]{llncs}
\usepackage[T1]{fontenc}
\usepackage{graphicx}
\usepackage{amsmath}
\usepackage{dsfont}
\usepackage{mathrsfs}
\usepackage{amssymb}
\usepackage{xcolor}
\usepackage{hyperref}
\begin{document}
% %
% \title{Population-based fMRI Classification with Contrastive Functional Connectivity Graph Learning}

\title{Contrastive Functional Connectivity Graph Learning for Population-based fMRI Classification}
\author{Xuesong Wang\inst{1} \and
Lina Yao\inst{1}\and
Islem Rekik\inst{2}\and
Yu Zhang\inst{3}}
% index{Wang, Xuesong}
% index{Yao, Lina}
% index{Rekik, Islem}
% index{Zhang, Yu}

%
\institute{University of New South Wales, NSW 2052, Australia \email{\{xuesong.wang1,lina.yao\}@unsw.edu.au}\and
Istanbul Technical University, Maslak 34469, Turkey
\\\and
Lehigh University, Bethlehem, PA 18015, USA\\
\email{yuzi20@lehigh.edu}}
\maketitle              % typeset the header of the contribution
\begin{abstract}
Contrastive self-supervised learning has recently benefited fMRI classification with inductive biases. Its weak label reliance prevents overfitting on small medical datasets and tackles the high intraclass variances. Nonetheless, existing contrastive methods generate resemblant pairs only on pixel-level features of 3D medical images, while the functional connectivity that reveals critical cognitive information is under-explored. Additionally, existing methods predict labels on individual contrastive representation without recognizing neighbouring information in the patient group, whereas interpatient contrast can act as a similarity measure suitable for population-based classification. We hereby proposed contrastive functional connectivity graph learning for population-based fMRI classification. Representations on the functional connectivity graphs are ``repelled'' for heterogeneous patient pairs meanwhile homogeneous pairs ``attract'' each other. Then a dynamic population graph that strengthens the connections between similar patients is updated for classification. Experiments on a multi-site dataset ADHD200 validate the superiority of the proposed method on various metrics. We initially visualize the population relationships and exploit potential subtypes. Our code is available at \url{https://github.com/xuesongwang/Contrastive-Functional-Connectivity-Graph-Learning} .

\end{abstract}
\section{Introduction}
Recently, self-supervised deep models such as contrastive learning have shown promising results in 3D medical image classification~\cite{Dufumier2021contrameta,xing2021contracategorical,zeng2021positional,azizi2021big}.  The pillar of contrastive learning is to augment a similar 3D image from one patient to form a ``homogeneous'' pair and use images from other patients to construct multiple ``heterogeneous'' pairs. The ``homo-'' contrasts are minimized while ``heter-'' contrasts are maximized. As it incorporates such inductive bias to generate pseudo-labels for augmented samples, contrastive learning enables training on small medical datasets in an unsupervised manner~\cite{li2021contrasmallscale}. Such weak reliance on ground truths prevents overfitting in large deep networks~\cite{zeng2021contrastive-overfitting}, and helps tackle high intraclass variances in brain disorder diseases induced by potential disease subtypes~\cite{liu2021metaconsite}.
%, inconsistent equipment meansurements on sites, missing personal charaterestic information%.

Existing contrastive learning frameworks process 3D fMRI images like typical 2D  datasets such as ImageNet~\cite{deng2009imagenet}, where they augment ``homo-'' pairs with transformations including random cropping, colour distortion, and downsampling~\cite{chen2020simclr,hu2021semi}. Functional Connectivity (FC), which quantifies the connection strengths between anatomical brain regions of interests (ROIs), may be neglected in those transformations and impair the brain disorder classification~\cite{mueller2013individual,rodriguez2019cognitive}, as they are related to cognitive behaviours focusing on perception and vigilance tasks~\cite{cohen2018behavioral}. Such negligence calls for a novel formalization of ``homo-'' and ``heter-'' pairs meanwhile preserving the FC information. Additionally, existing frameworks treat contrastive learning as a self-supervised encoder where a single patient embedding is an input to the classification network to obtain a single output~\cite{Dufumier2021contrameta,konkle2020instance}. They tend to overlook that interpatient contrast itself is a similarity measure, which can intrinsically enable a population-based classification. In the population graph, every patient can represent a node and the similarity between two patients forms a weighted edge~\cite{parisot2017spectral,bessadok2021graphsurvey}. Despite the fruitful benefits population-based classification brings to medical datasets, for instance, it alleviates high-intraclass variances by forming sub-clusters of the same class in the graph~\cite{zhang2022individual,zhong2021graphclustering}, and also stabilizes the prediction metrics% by referring to labelled neighours, similar to a bootstrapping strategy
~\cite{chen2021hierarchical}, population-based classification on contrastive features has not been fully explored. 

We aim to classify brain disorders on a population graph using contrastive FC features to tackle the aforementioned challenges. We primitively enable contrastive learning on FC by defining a ``homo-'' pair with two FC graphs generated from the non-overlapping ROI time series of the same patient, based on which a spectral graph network capable of convolving full ROI connections is optimized. The population-based classifier then adopts the contrastive embedding to dynamically the update population graph structure and utilize neighbouring information of each patient for prediction. Our major contributions are detailed as follows:
\begin{itemize} 
    \item Formalization of the ``homogeneous'' and ``heterogeneous'' pairs  which primitively enables contrastive learning optimization on FC graphs.%Spectral convolution networks are employed to present global FC connection information.
    \item Contrastive features for population-based classification. We utilize dynamic graph convolution networks to enhance patient clusters within the graph.
    \item Population graph visualization with the implication of subtypes. We initially visualize the ADHD population graph and reveal two potential subtypes for the health control group.
\end{itemize}

\section{Method}
\subsection{Problem Formalization.}
Given a dataset of \(P\) patients \(\{(\mathcal{G}_i^j, y_i)\}_{i=1}^P\) , where \(y_i \in \{0, 1\}\) represents the class for the \(i\)-th patient, \(\mathcal{G}_i^j = (\mathcal{V}_i^j, \mathcal{E}_i^j)\) is the \(j\)-th view of the individual FC graph where Pearson correlations for each ROI is calculated for node features \(\mathcal{V}_i^j\) and edges \(\mathcal{E}_i^j\) are constructed using partial correlations between ROIs. Multiple views are generated by slicing ROI timeseries into non-overlapping windows. As illustrated in Fig~\ref{Fig: framework}, Our objective is to construct a new population graph \(\mathcal{G^P}\) and build a model \(\textbf{\textit{h}}\) to predict node labels in  \(\mathcal{G^P}: \textbf{\textit{h}}(\mathcal{G^P}) \in \mathbb{R}^{P \times class}\). \( \mathcal{G^P} = (\mathcal{V^P}, \mathcal{E^P})\) is constructed with \(\mathcal{V^P} = \textbf{\textit{f}}(\mathcal{G}_{i=1, 2, \cdots, P})\in \mathbb{R}^{P \times d}\) that embeds each patient to a \(d\)-dimensional vector and \(\mathcal{E^P}\) represents the attractiveness between patients. Labels are provided for training nodes in \(\mathcal{G^P}\)  while the testing labels are masked.
 
\subsection{Overall Framework}

\begin{figure}
\centering
\includegraphics[width=0.95\textwidth]{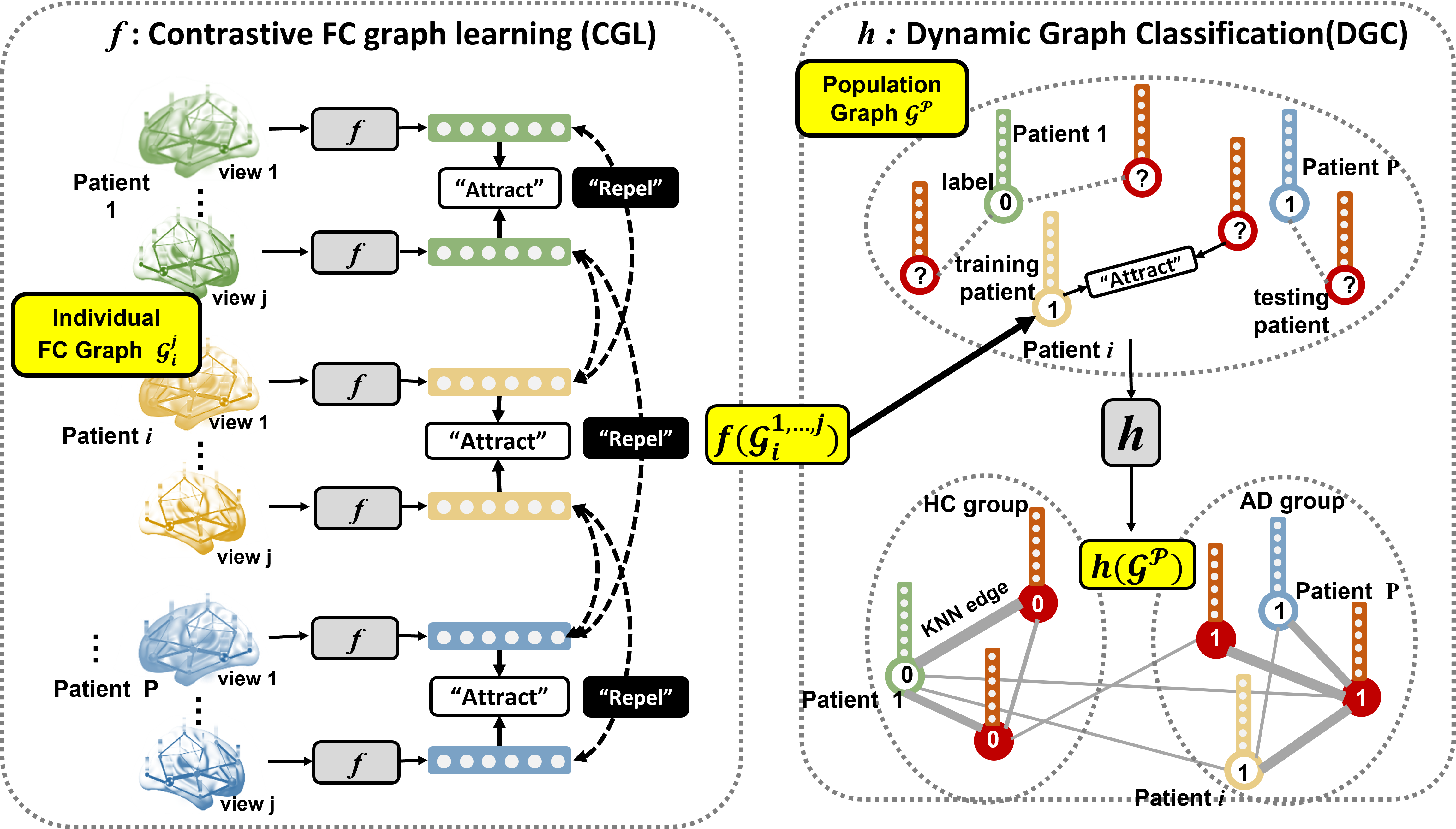}
\caption{The framework of our method. Contrastive FC graph learning (CGL) extracts contrastive embeddings from multiview of each patient, based on which a population-graph is constructed for dynamic graph classification (DGC).} \label{Fig: framework}
\end{figure}

The framework overflow is shown in Fig.~\ref{Fig: framework}. Multiple non-overlapping views of N of patients are initially sampled to obtain FC graph inputs. The motivation of contrastive learning is that two inputs coming from the same patient should form a ``homo-'' pair and ``attract'' each other; otherwise, two from different patients should form a ``heter-'' pair and ``repel''. A spectral graph convolutional network \(\textbf{\textit{f}}\) is then trained in a self-supervised manner to acquire contrastive embedding of each patient. A population graph is further defined by a collection of patient contrastive embeddings, based on which a dynamic graph classifier is trained to predict a label for each patient node. It applies a dynamic edge convolution network \(\textbf{\textit{h}}\) to node features. Since K-nearest neighbours establish new edges with adaptive node features, the population graph structure is dynamically modelled to strengthen connections between ``attracted'' patients.

\subsection{\underline{C}ontrastive Functional Connectivity \underline{G}raph \underline{L}earning (CGL).}

We propose contrastive FC learning to train deep networks on small medical datasets in a self-supervised manner, while preserving FC node-edge relationships. FC graph input of the \(i\)-th patient is represented by a functional connectivity graph \(\mathcal{G}_i = (\mathcal{V}_i, \mathcal{E}_i)\). Pearson correlations between ROIs are used for node features, where the \(r\)-th ROI is represented by \(v^{(r)}_i %\in \mathcal{V}_i , v^{(r)}_I 
\in \mathbb{R}^{ROIs}\) indicating Person coefficients of all connecting ROIs. Edges are constructed using partial correlations to create sparser connections and simplify the graph structure. The module is detailed as follows:

\subsubsection{Spectral Graph Convolution \(\textbf{\textit{f}}\) on FC graphs.}
Unlike conventional graph convolutions that pass messages to a single ROI node from its adjacency, spectral graph networks simultaneously convolve full node connections by incorporating the Laplacian matrix \(\mathcal{L} =  I - D^{-1/2}AD^{-1/2}\), where \(I\) is the identity matrix, \(A \in \mathcal{R}^{ROIs \times ROIs}\) is the adjacency matrix and \(D  \in \mathcal{R}^{ROIs \times ROIs} \) refers to the diagonal degree matrix. This matches with the domain knowledge about brain disorders that certain isolated FCs cooperate to achieve a cognitive task~\cite{cohen2018behavioral}, therefore incorporating all node features will induce a robust representation. A spectral graph convolution block is defined as:
\begin{equation}
\label{Eq: spectralgcn}
   \mathcal{V}'_i = \sum_{k=1}^{K} Z^{(k)}(\mathcal{V}_i) \cdot\theta^{(k)}
\end{equation}
 \noindent where \(k\) is the filter size, \(Z^{(k)}(\mathcal{V}_i)\) is recursively computed as: 
\(Z^{(k)}(\mathcal{V}_i)  = 2 \tilde{\mathcal{L}} Z^{(k-1)}\cdot\)
\\
\((\mathcal{V}_i)- Z^{(k-2)}(\mathcal{V}_i), \quad  \text{with} Z^{(1)}(\mathcal{V}_i)  = \mathcal{V}_i , Z^{(2)}(\mathcal{V}_i)  = \tilde{\mathcal{L}} \mathcal{V}_i\). \(\tilde{\mathcal{L}}\) is the scaled and normalized Laplacian 
\(\tilde{\mathcal{L}} = \frac{2\mathcal{L}}{\lambda_{max}} - I\) with \(\lambda_{max}\) denoting the largest eigenvalue of \(\mathcal{L}\). \(\theta^{(k)}\) are learnable parameters. In practice, K = 3, hence every spectral convolutional block includes a single node projection \(\mathcal{V}_i \theta^{(1)}\), all nodes projection \( \tilde{\mathcal{L}} \mathcal{V}_i \theta^{(2)}\) and a combination of both with \(Z^{(3)}(\mathcal{V}_i)\theta^{(3)}\).
To prevent overfitting, the CGL module \(\textbf{\textit{f}}\) is comprised of 2 spectral graph convolution blocks with top-K graph pooling blocks. Each top-K graph pooling deactivates a subset of nodes to form a smaller graph. The resulting embedding is optimized using contrastive loss and then input to the classification.

% \begin{equation}
% \label{Eq: top-k pooling}
%   p_i = \frac{\mathcal{V}_i W}{\parallel W\parallel}, \quad  \mathcal{V}'_i = (\mathcal{V}_i \odot \tanh{(p_i)})_{top_K}, \quad \mathcal{E}'_i = \mathcal{E}_i(\mathcal{V}'_i,  \cdot)
% \end{equation}
% \noindent where \(W\) is a gate projection to evaluate the drop-out probability of certain nodes, based on which we keep the top-K nodes as well as their connections. In practice, activated nodes are halved after each pooling.

\subsubsection{Contrastive Optimization.}
After the spectral graph convolutions \(\textbf{\textit{f}}\), each view of a patient is embedded as:  \(\textbf{\textit{v}}_i^j = \textbf{\textit{f}}(\mathcal{G}_i^j)\). The motivation of contrastive optimization is to enforce attractions among embeddings that are resemblant without referring to the ground truths. For a batch of N patients with 2 views per patient, we can construct an attraction matrix \(\mathcal{M} \in 2N \times 2N\) where its element is a similarity measure between a pair of random two views:

\begin{equation}
\label{Eq: similarity measure}
  \mathcal{M}(m, n) = \frac{\textbf{\textit{v}}_m^\top \textbf{\textit{v}}_n}{\parallel \textbf{\textit{v}}_m\parallel  \parallel \textbf{\textit{v}}_n\parallel}
\end{equation}

We define a ``homogeneous'' pair as two embeddings \(v_m\) and \(v_n\) coming from the same patient. As they are generated from two non-overlapping ROI time series, their original FC graph inputs can be diverse. However, semantically they are representing the same patient hence should have similar embeddings. The contrastive loss function on this ``homo-'' pair is defined as:
\begin{equation}
\label{Eq: paired contra loss function}
  \mathscr{L}(m, n) = - \log \frac{\exp{(\mathcal{M}(m, n)/\tau))}}{\sum_{i=1}^{2N}\mathds{1}_{i \neq m}\exp{(\mathcal{M}(m, i)/\tau)}}
\end{equation}
\noindent where \(\tau\) is a temperature factor to control the desired attractiveness strength. \(\mathds{1}(\cdot) = \{0, 1\}\) is an indicator function that zeros self-attraction \(\mathcal{M}(m, m)\). During training, the numerator is maximized by increasing the ``homo-'' attraction and the denominator is minimised with ``heter-'' pairs ``repeling'' each other. The loss function is not commutative as the denominator will compute ``heter-'' pairs differently. The overall contrastive loss is the sum of \(2N\) ``homo-'' pairs in the batch: \(L_{CGL} = \frac{1}{2N}\sum_{i=1}^{N}(\mathscr{L}(2i-1, 2i) + \mathscr{L}(2i, 2i-1))\).
The contrastive loss function is different from typical similarity loss and softmax loss because it simultaneously optimizes ``Multiple-Homo-Multiple-Heter'' pairs rather than ``Single-Homo-Single-Heter'' pairs (similarity loss) and ``Single-Homo-Multiple-Heter'' pairs (softmax loss). This avoids bias towards a single pair, hence stabilizing the representation.

\subsection{Population-based \underline{D}ynamic \underline{G}raph \underline{C}lassification (DGC).}
Now that we have the embedding \(\textbf{\textit{v}}_i^j = \textbf{\textit{f}}(\mathcal{G}_i^j)\) that can measure patient similarity, the motivation is to build an attraction graph \(\mathcal{G}^P\) for population-based classification. When predicting the label for an unknown patient \(v^P_i\) in the graph, the model can infer from its labelled neighbours. In order to construct a graph with an adaptive structure, we propose dynamic edge convolution for classification.

\subsubsection{Dynamic Edge Convolution \(\textbf{\textit{h}}\) on the Population Graph.} Dynamic edge convolutions form an initial graph with isolated patient nodes, then project node features to a new space and dynamically construct new edges on the new feature space with top-K connection strengths  defined as:
\begin{equation}
    v_i^{P'} = \sum_{m\in \mathcal{E}^ P_{(i, \cdot)}} {\phi (v_i^{P} \parallel v^{P}_m - v^{P}_i)} \quad,  \quad  \mathcal{E}^P = KNN(\mathcal{V}^P)_{topK}   
\end{equation}

\noindent where \(\parallel\) is the concatenation function, \(\phi(\cdot)\) is a fully connected network. Initially, the edge-convolution layer \(\phi\) projects the node features \(v_i^{P}\) as well as its connecting edge features\((v^{P}_m - v^{P}_i)\) to update node features. New edges are built by selecting top-K similarities. For the next update, connections between closer patients are strengthened while edges on distant patients are abandoned. The final classifier layer is a fully connected network to predict each node's class probability \(Pr\) where training nodes are labelled. We adopted focal loss for the classification: \(L_{DGC} = - (1 - Pr)^\gamma\log(Pr)\).
As suggested by~\cite{zhao2022dynamic}, the focal loss lowers the threshold of accepting a class, inducing a higher recall rate for disorder groups. As KNN is a non-parametric model, trainable parameters come only from node projection, narrowing the parameter searching space and preventing overfitting on the population graph.

\section{Experiments}
\subsection{Dataset and experimental details.}
A multi-site fMRI dataset ADHD200 \footnote{http://preprocessed-connectomes-project.org/adhd200/} is used for comparing the proposed method with baseline models. We collect 596 patients data on AAL90 ROIs to construct individual FC graphs. As implied in ~\cite{zhao2022dynamic}, incorporating personal characteristic data (PCD) such as age helps to stabilize the metrics. We accordingly keep the three sites KU, KKI, and NYU with no missing values on the 7 PCD features:  age, gender, handedness, IQ Measure, Verbal IQ, Performance IQ, and Full4 IQ. Data from the rest sites are excluded due to the lack of PCD information or few ADHD samples. Patients are sampled within each site and combined afterwards. The ratio of site training/validating/testing split is 7: 1: 2, and we select patients proportionally to the class-imbalance ratio.
 
We compare four baseline models and two variants of our methods on four metrics: AUC, accuracy, sensitivity, and specificity. KNN is trained with raw vectorized FC features. MLP utilizes two views of FC vectors for classification~\cite{chen2019multichannel}. BrainGNN~\cite{li2021braingnn} adopts the same FC graph as ours and introduces a novel GNN structure. A population-based method SpectralGCN~\cite{parisot2017spectral} uses supervised learning to train a spectral graph convolution for similarity measures. For ablation studies, we test dynamic graph classification on a population graph using raw FC features (DGC) and perform contrastive graph learning (CGL) with a KNN classifier to enable unsupervised learning. Regarding implementation details,  we run the model with a batch size of 100 for 150 epochs. 
Adam optimizer is selected, and
learning rates for CGL and DGC are 0.001 and 0.005, respectively.%The learning rates are reduced by half every 20 epochs for better convergence. 
PCD features are concatenated to each ROI feature. temperature factor \(\tau\) in contrastive loss function is 0.1. %We adopt the same pipeline of the patient graph encoder \textbf{\textit{f}} as ~\cite{li2021braingnn} 
The filter size for spectral convolution is 3. For dynamic edge convolution, the structure resembles~\cite{wang2019dynamic} where 20 nearest neighbours are used for edges. All methods with implemented with PyTorch and trained with GPU TITAN RTX.

\subsection{Classification results.}

Table~\ref{tab1: classification metrics} shows the four metrics of the comparing methods on the three sites. Each method is tested for five runs with the mean and standard values. Our method significantly outperforms baselines on AUC and achieves the best average accuracy. Due to class imbalance in site KKI with fewer ADHD samples, baseline methods tend to have larger variations on sensitivity and specificity. For example, BrainGNN and CGL achieve higher sensitivity and specificity in the sacrifice of the other metric, whereas our method achieves relatively stabilized and balanced metrics. It is worth mentioning that our variant CGL achieves promising AUC and ACC as a completely self-supervised model without reliance on labels, manifesting the effectiveness of the contrastive training framework. The introduction of focal loss to our framework balances the metrics and increases the recall rate of ADHD, which is medically significant.

\begin{table}[]
\caption{Classification results on 3 sites (PKU, KKI, NYU) of ADHD dataset.\\
\centering
DGC and CGL are two variants of the proposed model. (\(\ast\): p < 0.05)}\label{tab1: classification metrics}
% \centering
\begin{tabular}{|c|c|c|c|c|}
\hline
Models      & AUC             & ACC (\%)    & SEN (\%)      & SPEC (\%)    \\
\hline
KNN~\cite{eslami2018KNN}     & 0.6816 \(\pm\) 0.0305 *  & 63.43 \(\pm\) 3.42 & 56.67 \(\pm\) 9.94    & 69.59  \(\pm\)4.86  \\
MLP~\cite{chen2019multichannel}         & 0.6266 \(\pm\) 0.0190 * & 55.43 \(\pm\) 3.90 * & 52.54 \(\pm\) 27.09  & 58.08 \(\pm\) 26.27  \\
BrainGNN~\cite{li2021braingnn}    & 0.6410 \(\pm\) 0.0659 * & 62.00 \(\pm\) 6.20 &  73.43 \(\pm\) 13.26  & 51.51 \(\pm\) 23.77  \\
SpectralGCN~\cite{parisot2017spectral} & 0.6321  \(\pm\) 0.0199 * & 59.00 \(\pm\) 3.05 * & 61.19 \(\pm\)  12.63 & 56.99 \(\pm\) 11.18 * \\
\hline
DGC (Var-1)  & 0.6647 \(\pm\)  0.0328 * & 61.00 \(\pm\) 3.02  & 58.21  \(\pm\) 6.47   & 63.56   \(\pm\) 6.80 \\
CGL (Var-2) & 0.7101 \(\pm\) 0.0306  & 66.71 \(\pm\) 3.08 & 53.13 \(\pm\) 4.59  *& {\bfseries 79.18  \(\pm\) 3.05} \\ 
\hline
{\bfseries CGL+DGC(ours)}    & {\bfseries 0.7210 \(\pm\) 0.0263}  & {\bfseries 67.00  \(\pm\) 3.79 }& 61.49  \(\pm\) 5.93  & 72.10 \(\pm\) 9.86\\
\hline
\end{tabular}
\end{table}

\subsection{Discussion.}
\subsubsection{Contrastive FC Graph Learning.}
\begin{figure}
\centering
\includegraphics[width=0.45\textwidth]{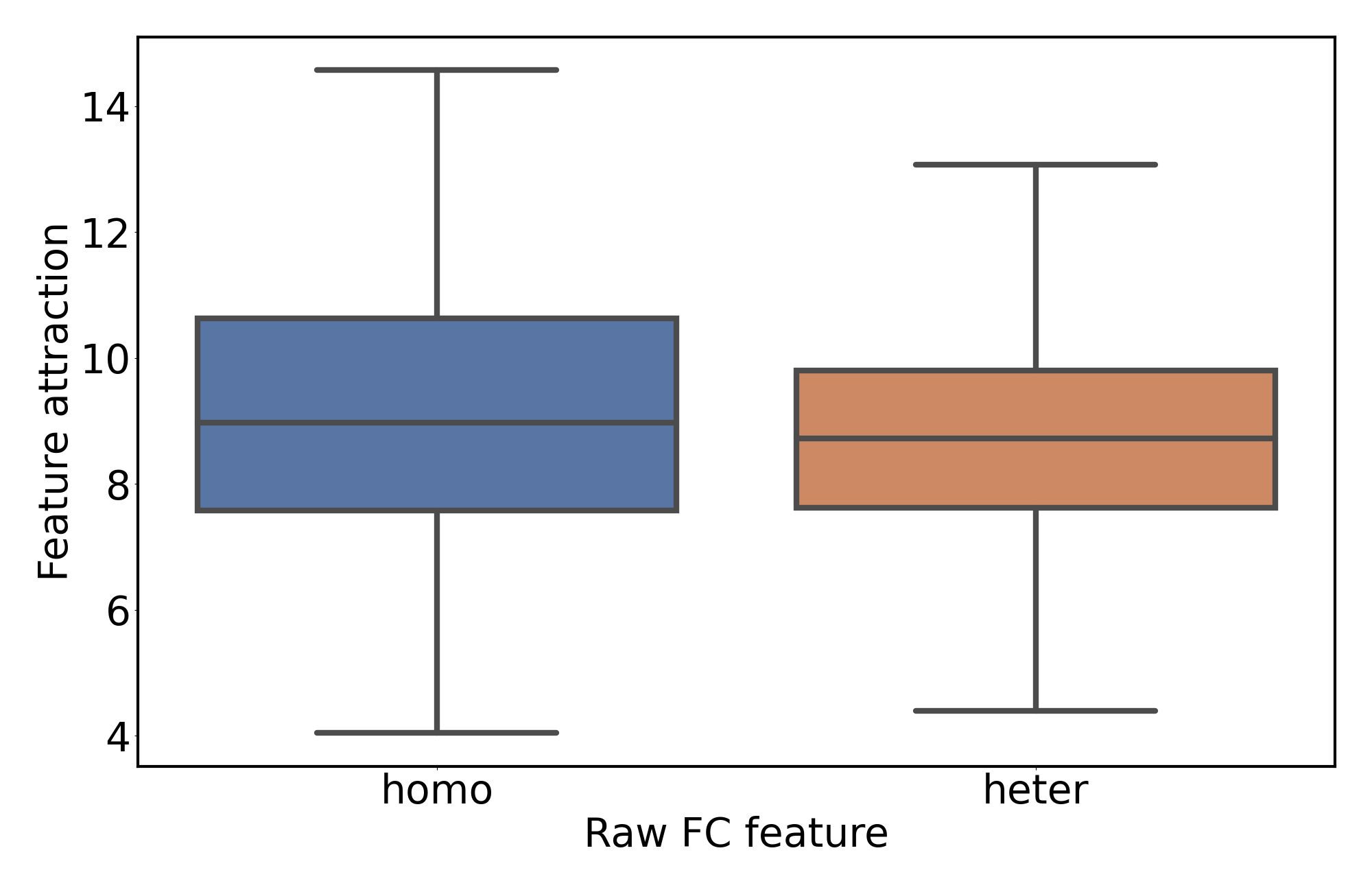}
\includegraphics[width=0.45\textwidth]{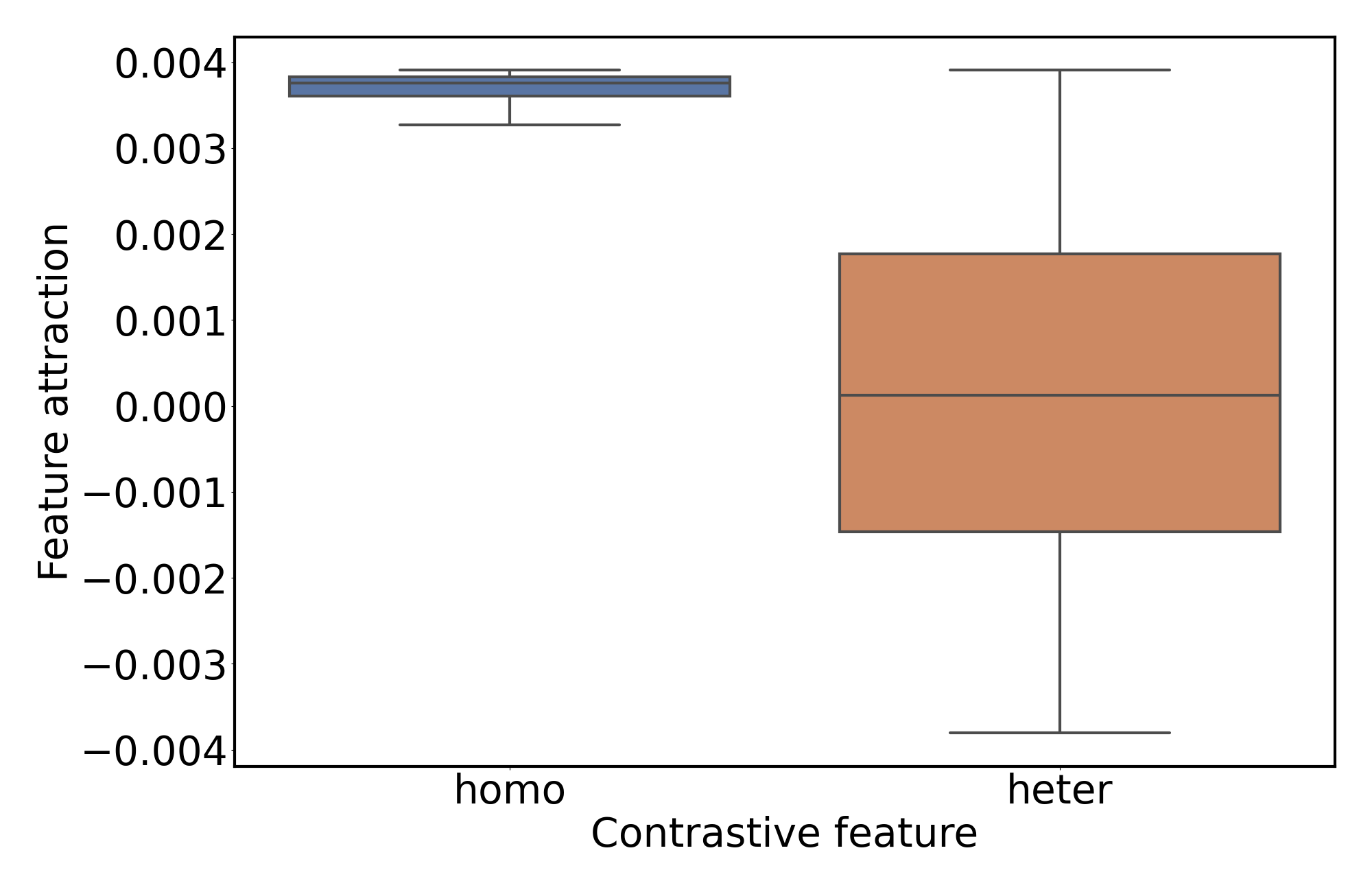}
\includegraphics[width=0.45\textwidth]{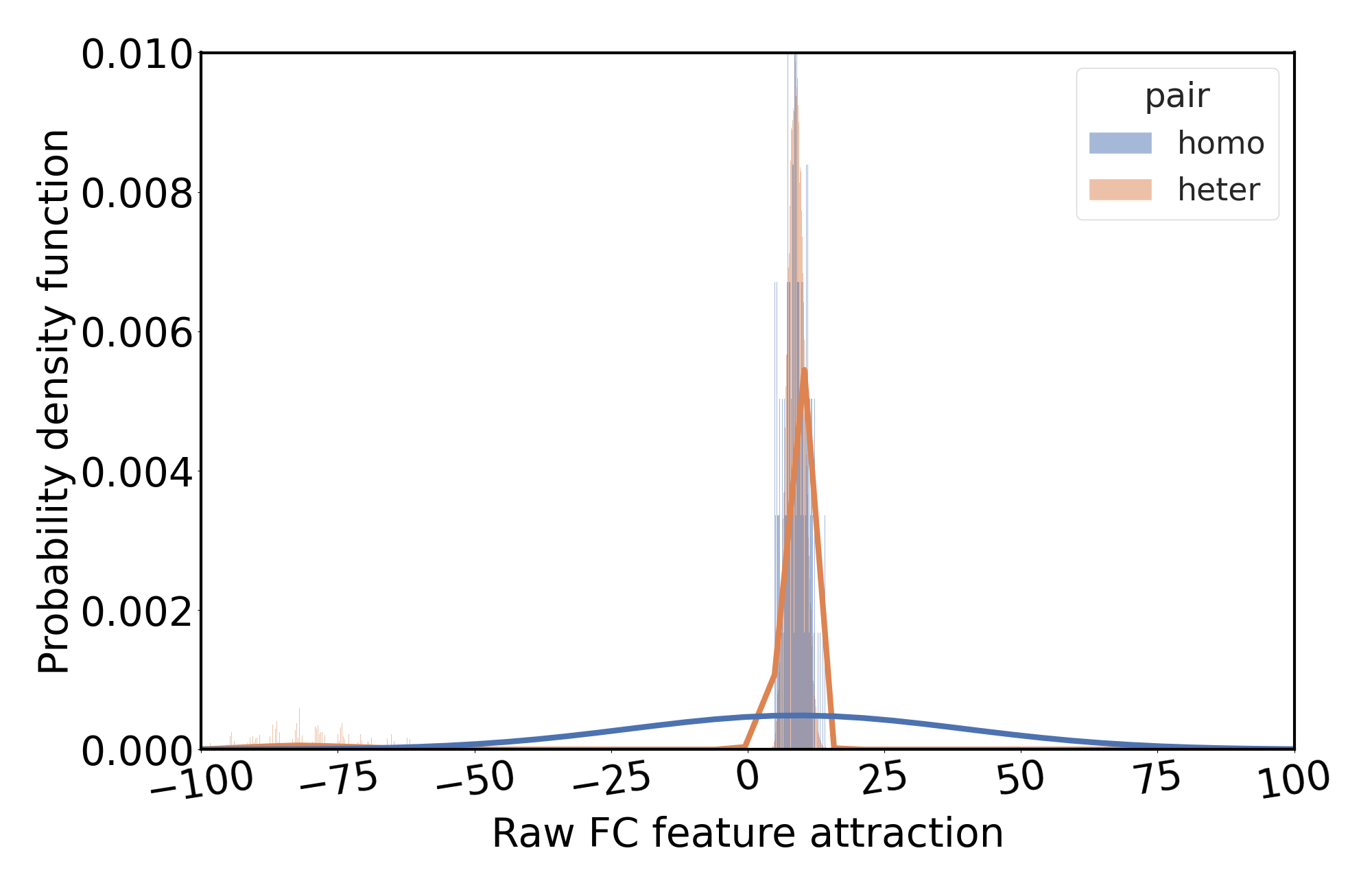}
\includegraphics[width=0.45\textwidth]{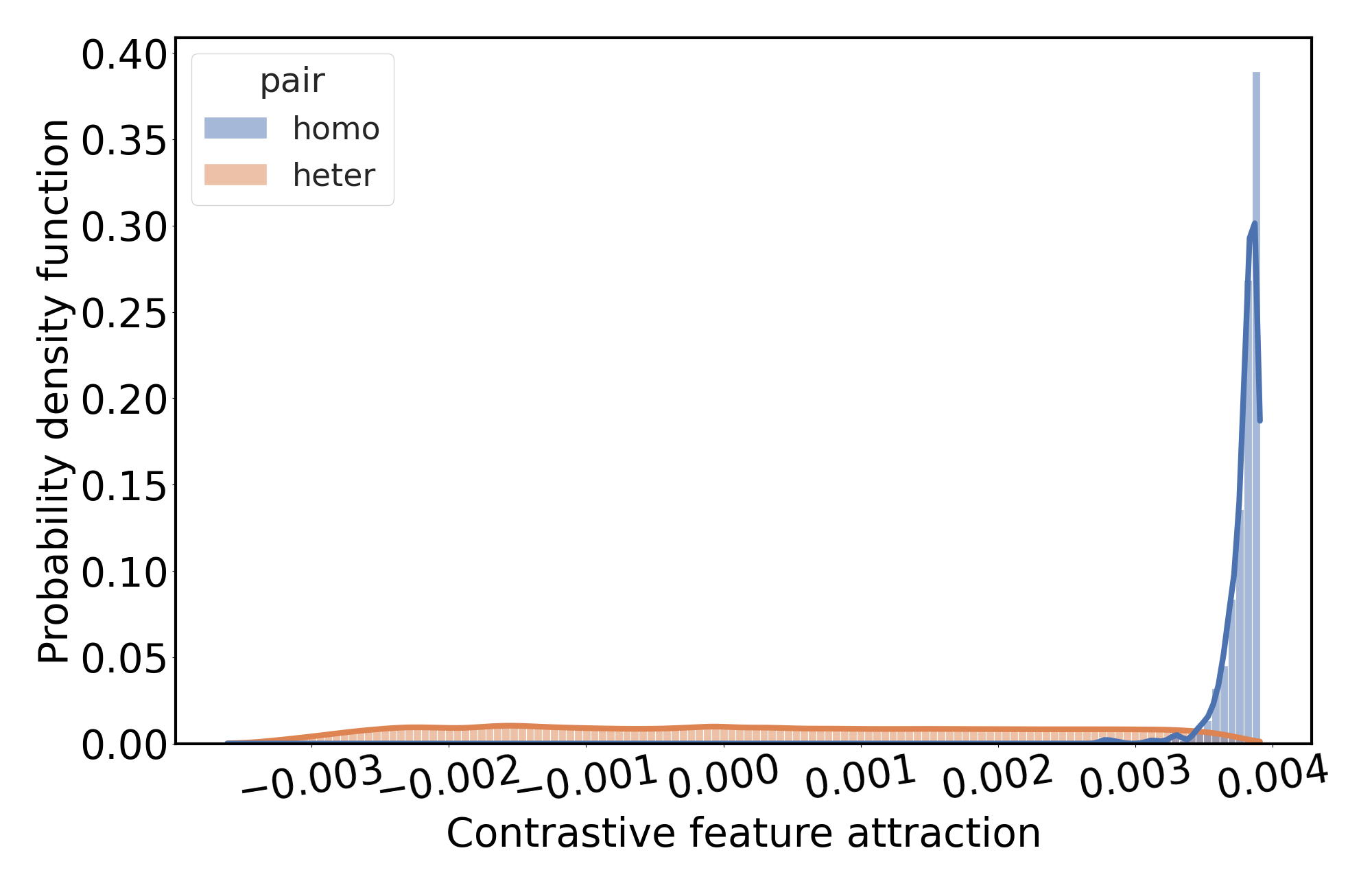}
\caption{Boxplot and probability density function of attractions on raw FC and contrastive features.} \label{Fig: violinplot}
\end{figure}

To verify the effectiveness of the contrastive FC graph learning, we aim to compare the patient attraction. The distributional similarity of ``homo-'' and ``heter-'' pairs is compared in Fig.~\ref{Fig: violinplot} on raw vectorized FC features and contrastive features. The results on the raw features group show no substantial differences, indicating equal similarity between homo- and heter-pairs. After the contrastive representation, similarities on the homo-pairs are significantly higher with tighter variance, meaning that the loss function managed to enforce homo-attraction. Meanwhile, the attractions on the heter-pairs are averaged around zero with large variance, suggesting that some heter-views can still be similar. Those similar heter-pairs can form the population graph.

\subsubsection{Population graph visualization.}
\begin{figure}
\centering
\includegraphics[width=0.32\textwidth]{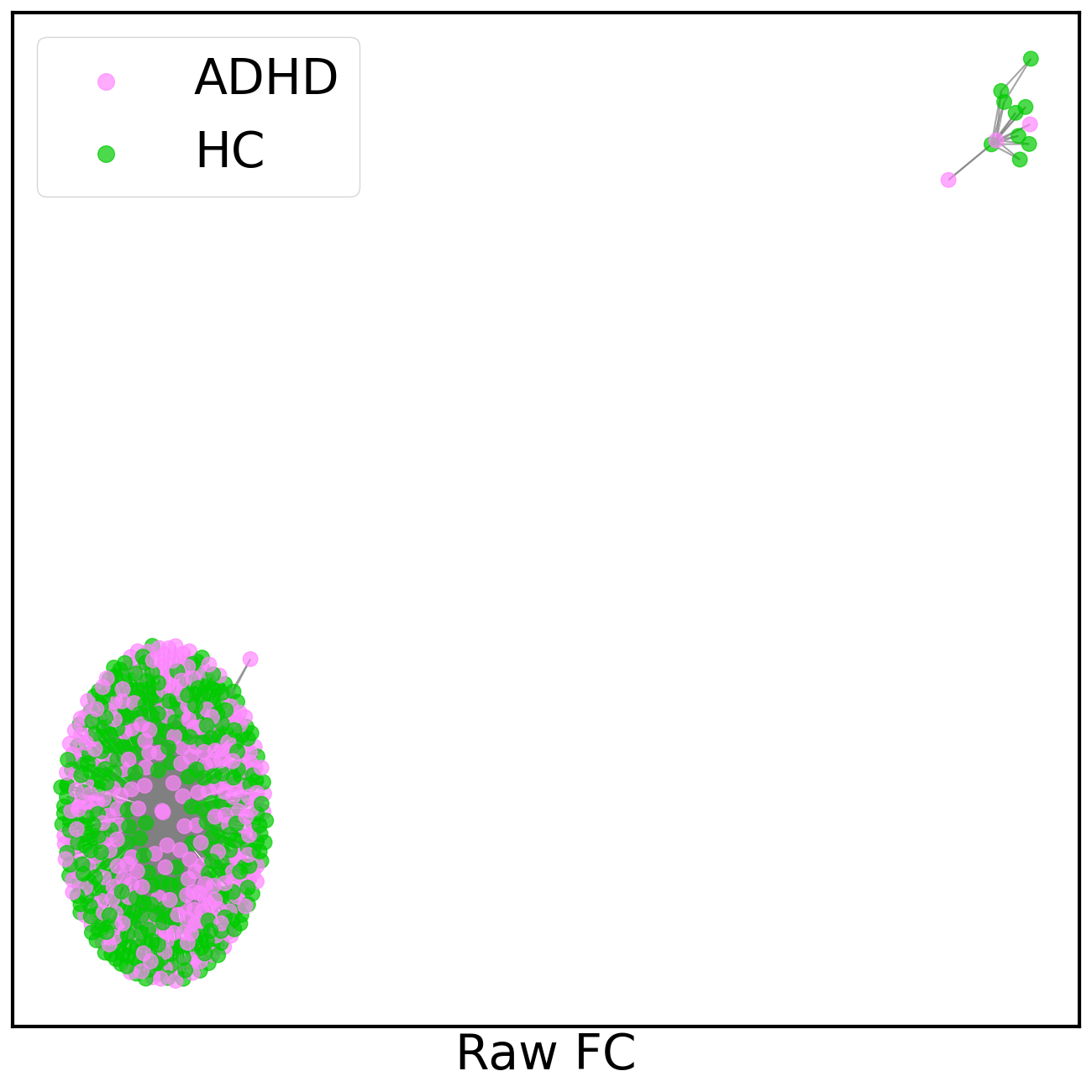}
\includegraphics[width=0.32\textwidth]{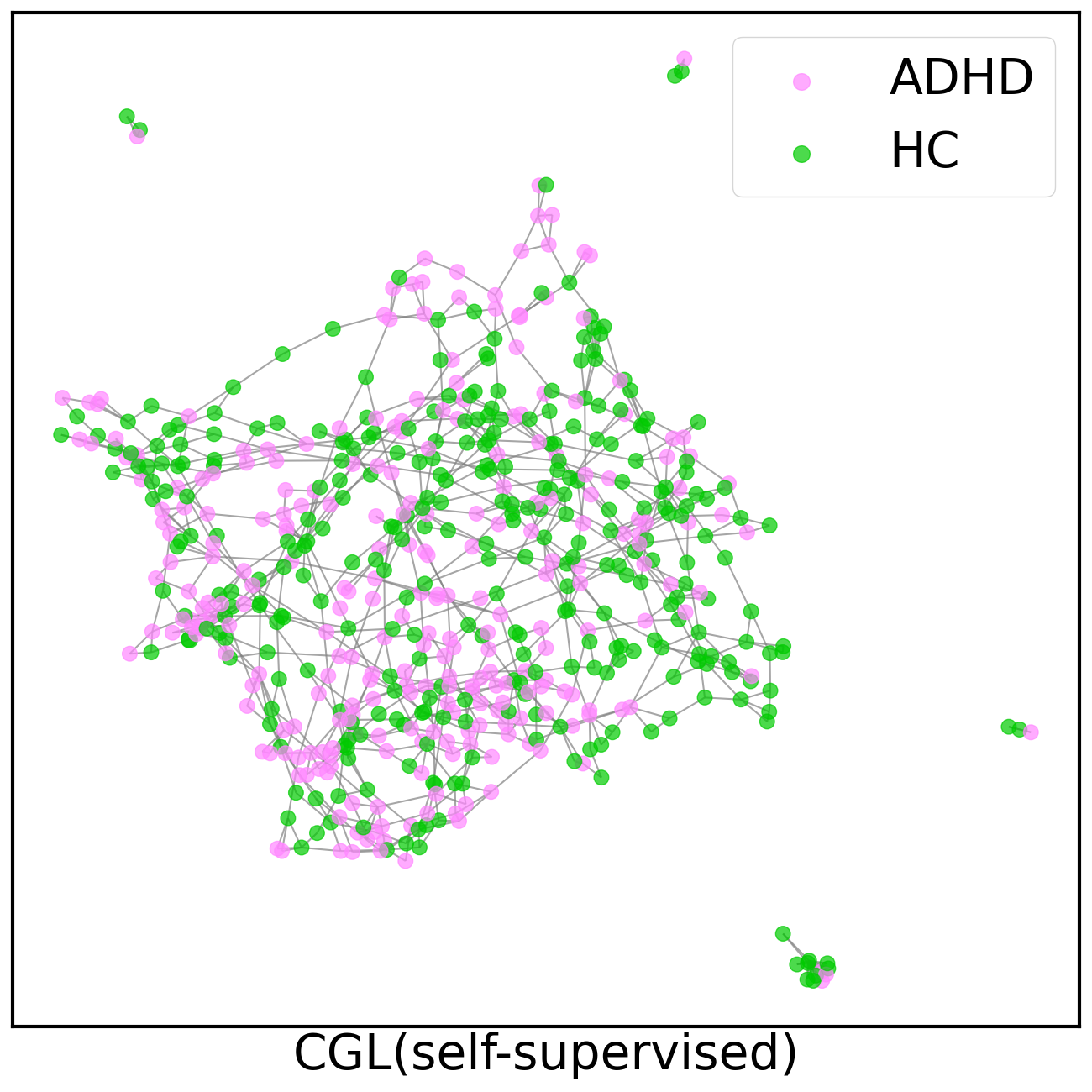}
\includegraphics[width=0.32\textwidth]{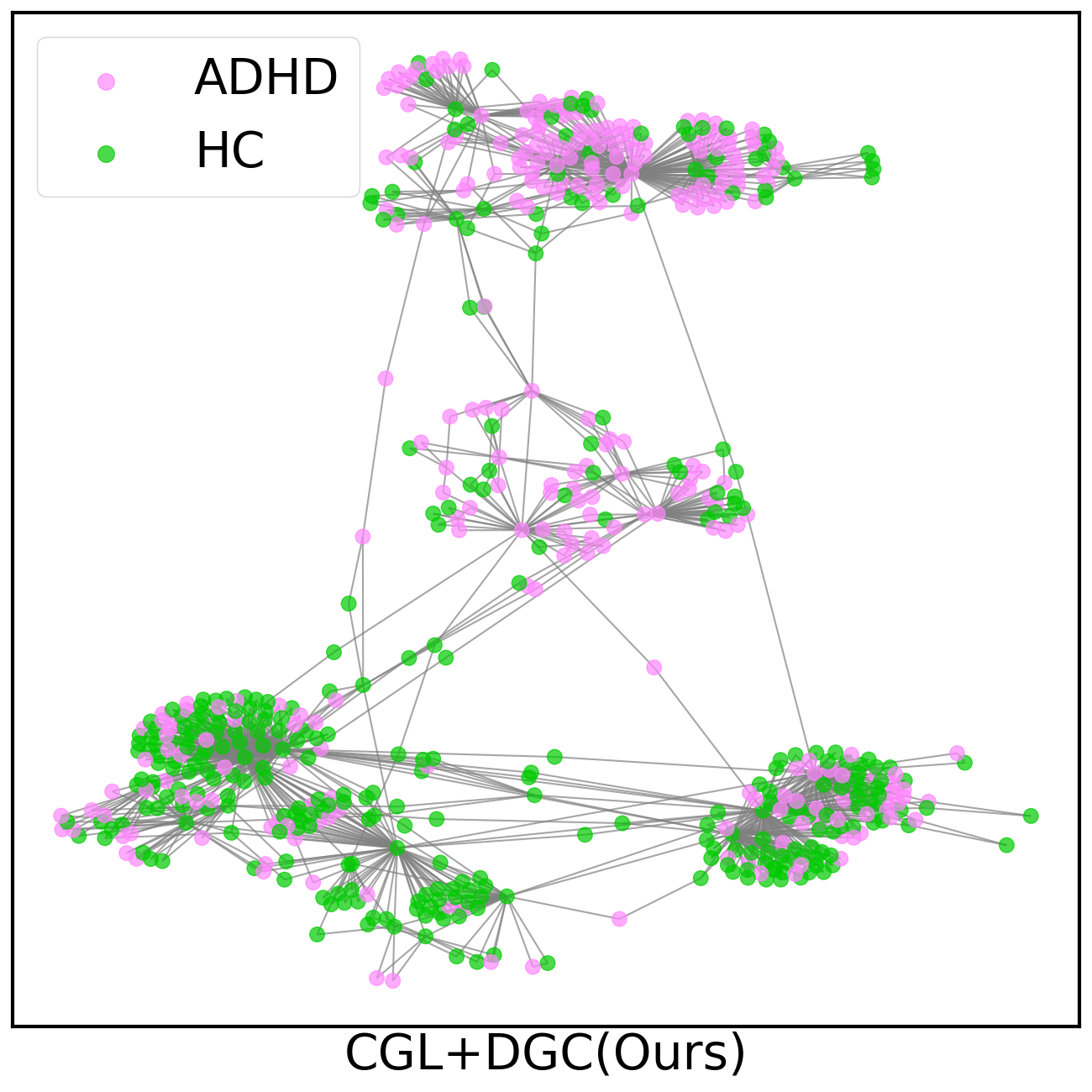}
\caption{Population graph visualization on three features.} \label{Fig: population graph}
\end{figure}

We construct the population graph with edges representing the patient similarities to unveil hidden patient relationships. Only two edges are visualized per patient to construct a cleaner graph. The results of raw FC vectors, self-supervised contrastive features, and the last hidden layer of our method are illustrated in Fig~\ref{Fig: population graph}. Two classes are denoted with different colours. 
Both classes are cluttered with raw FC features, whereas contrastive features reveal limited patterns with self-supervised learning; for instance, the lower-left tends to have more ADHD patients connecting each other while more HC patients are clustered on the right. Our method presents the most distinct patterns with four clear clusters, each formed by multiple edges sharing mutual nodes, i.e., representative patients. It also suggests two subtypes in HC, potentially leading to intraclass variance and deteriorating the metrics. We will leave the exploration and validation of subtypes for future work.

\section{Conclusion}
We have addressed two critical questions in fMRI classification: Can contrastive learning work on Functional Connectivity graphs? Can contrastive embeddings benefit population-based classification? We hereby define a ``homogeneous'' pair in the contrastive loss as two FC graphs generated from non-overlapping ROI time series of the same patient. A ``heterogeneous'' pair is formalized with two FC graphs from different patients. A spectral graph network capable of full FC connection convolution is optimized on multiple-homo-multiple heter pairs for contrastive learning. The resulting contrastive embeddings on an ADHD200 dataset show that stronger attractions are enforced on homo-pairs, indicating a similarity measure that enables a population graph classification. We then employ a dynamic edge convolution classifier to utilize such similarities in the population graph. Our dynamic graph classification achieves significant improvements on AUC and the highest average accuracy, meanwhile balancing sensitivity and specificity. Our visualization of the resulting population dynamic graph structure originally implies two potential subtypes of the health control group.

\bibliographystyle{splncs04}
\bibliography{bio}

% \clearpage
% \appendix
% \section{Model details.}

% \begin{figure}
% \centering
% \includegraphics[width=0.94\textwidth]{figs/CGL_workflow.png}
% \caption{Contrastive Graph Learning(CGL) workflow.} \label{Fig:CGL workflow}
% \end{figure}

% \begin{figure}
% \centering
% \includegraphics[width=0.94\textwidth]{figs/DGC_workflow.png}
% \caption{Dynamic Graph Classification(DGC) workflow.} \label{Fig: DGC workflow}
% \end{figure}

% \section{Additional results.}
% \begin{figure}
% \centering
% \includegraphics[width=0.32\textwidth]{figs/metric_plot_PK.png}
% \includegraphics[width=0.32\textwidth]{figs/metric_plot_KKI.png}
% \includegraphics[width=0.32\textwidth]{figs/metric_plot_NYU.png}
% \caption{Four metrics on three individual sites: PK, KKI, NYU.} \label{Fig: individual sites}
% \end{figure}

% \begin{figure}
% \centering
% \includegraphics[width=0.4\textwidth]{figs/MLP.png}
% \includegraphics[width=0.4\textwidth]{figs/BrainGNN.png}
% \includegraphics[width=0.4\textwidth]{figs/SpectralGCN.png}
% \includegraphics[width=0.4\textwidth]{figs/DGC.png}
% \caption{Population graph visualization on four baselines.} \label{Fig: baseline population graph}
% \end{figure}

\end{document}